%% file: template.tex
\documentclass{INTERSPEECH2023}

\interspeechcameraready

\title{Integrating Pretrained ASR and LM to Perform Sequence Generation for Spoken Language Understanding}
\name{Siddhant Arora$^1$, Hayato Futami$^2$, Yosuke Kashiwagi$^2$,\\ Emiru Tsunoo$^2$, Brian Yan$^1$, Shinji Watanabe$^1$}
\address{
  $^1$Carnegie Mellon University, U.S.A.\\
  $^2$Sony Group Corporation, Japan}
\email{siddhana@andrew.cmu.edu}
\newcommand{\Sref}[1]{\S\ref{#1}}
\usepackage[
backend=biber,
style=ieee,
citestyle=numeric-comp,
maxbibnames=3,
maxcitenames=3,
doi=false,isbn=false,url=false,eprint=false
]{biblatex}
\usepackage{booktabs}
\usepackage{multirow}
\usepackage{graphicx}
\usepackage{subcaption}
\defbibheading{bibliography}[\refname]{}

\DeclareSourcemap{
	\maps[datatype=bibtex, overwrite=true]{
		\map{
			\step[fieldsource=booktitle,
			match=\regexp{.*Interspeech.*},
			replace={Proc. Interspeech}]
			\step[fieldsource=journal,
			match=\regexp{.*INTERSPEECH.*},
			replace={Proc. Interspeech}]
			\step[fieldsource=booktitle,
			match=\regexp{.*ICASSP.*},
			replace={Proc. ICASSP}]
			\step[fieldsource=booktitle,
			match=\regexp{.*icassp_inpress.*},
			replace={Proc. ICASSP (in press)}]
			\step[fieldsource=booktitle,
			match=\regexp{.*Acoustics,.*Speech.*and.*Signal.*Processing.*},
			replace={Proc. ICASSP}]
			\step[fieldsource=booktitle,
			match=\regexp{.*International.*Conference.*on.*Learning.*Representations.*},
			replace={Proc. ICLR}]
			\step[fieldsource=booktitle,
			match=\regexp{.*International.*Conference.*on.*Computational.*Linguistics.*},
			replace={Proc. COLING}]
			\step[fieldsource=booktitle,
			match=\regexp{.*SIGdial.*Meeting.*on.*Discourse.*and.*Dialogue.*},
			replace={Proc. SIGDIAL}]
			\step[fieldsource=booktitle,
			match=\regexp{.*International.*Conference.*on.*Machine.*Learning.*},
			replace={Proc. ICML}]
			\step[fieldsource=booktitle,
			match=\regexp{.*North.*American.*Chapter.*of.*the.*Association.*for.*Computational.*Linguistics:.*Human.*Language.*Technologies.*},
			replace={Proc. NAACL}]
			\step[fieldsource=booktitle,
			match=\regexp{.*Empirical.*Methods.*in.*Natural.*Language.*Processing.*},
			replace={Proc. EMNLP}]
			\step[fieldsource=booktitle,
			match=\regexp{.*Association.*for.*Computational.*Linguistics.*},
			replace={Proc. ACL}]
			\step[fieldsource=booktitle,
			match=\regexp{.*Automatic.*Speech.*Recognition.*and.*Understanding.*},
			replace={Proc. ASRU}]
			\step[fieldsource=booktitle,
			match=\regexp{.*Spoken.*Language.*Technology.*},
			replace={Proc. SLT}]
			\step[fieldsource=booktitle,
			match=\regexp{.*Speech.*Synthesis.*Workshop.*},
			replace={Proc. SSW}]
			\step[fieldsource=booktitle,
			match=\regexp{.*workshop.*on.*speech.*synthesis.*},
			replace={Proc. SSW}]
			\step[fieldsource=booktitle,
			match=\regexp{.*Advances.*in.*neural.*information.*processing.*},
			replace={Proc. NeurIPS}]
			\step[fieldsource=booktitle,
			match=\regexp{.*Advances.*in.*Neural.*Information.*Processing.*},
			replace={Proc. NeurIPS}]
			\step[fieldsource=booktitle,
			match=\regexp{.*Workshop.*on.* Applications.* of.* Signal.*Processing.*to.*Audio.*and.*Acoustics.*},
			replace={Proc. WASPAA}]
			\step[fieldsource=publisher,
			match=\regexp{.+},
			replace={{}}]
			\step[fieldsource=month,
			match=\regexp{.+},
			replace={{}}]
			\step[fieldsource=location,
			match=\regexp{.+},
			replace={{}}]
			\step[fieldsource=address,
			match=\regexp{.+},
			replace={{}}]
			\step[fieldsource=organization,
			match=\regexp{.+},
			replace={{}}]
		}
	}
}
\begin{document}

\maketitle
 
\begin{abstract}
There has been an increased interest in the integration of pretrained speech recognition (ASR) and language models (LM) into the SLU framework. However, prior methods often struggle with a vocabulary mismatch between pretrained models, and LM cannot be directly utilized as they diverge from its NLU formulation. In this study, we propose a three-pass end-to-end (E2E) SLU system that effectively integrates ASR and LM subnetworks into the SLU formulation for sequence generation tasks. In the first pass, our architecture predicts ASR transcripts using the ASR subnetwork. This is followed by the LM subnetwork, which makes an initial SLU prediction. Finally, in the third pass, the deliberation subnetwork conditions on representations from the ASR and LM subnetworks to make the final prediction. Our proposed three-pass SLU system shows improved performance over cascaded and E2E SLU models on two benchmark SLU datasets, SLURP and SLUE, especially on acoustically challenging utterances.

\end{abstract}

\begin{figure*}[t]
\centering
\includegraphics[width=0.82\textwidth]{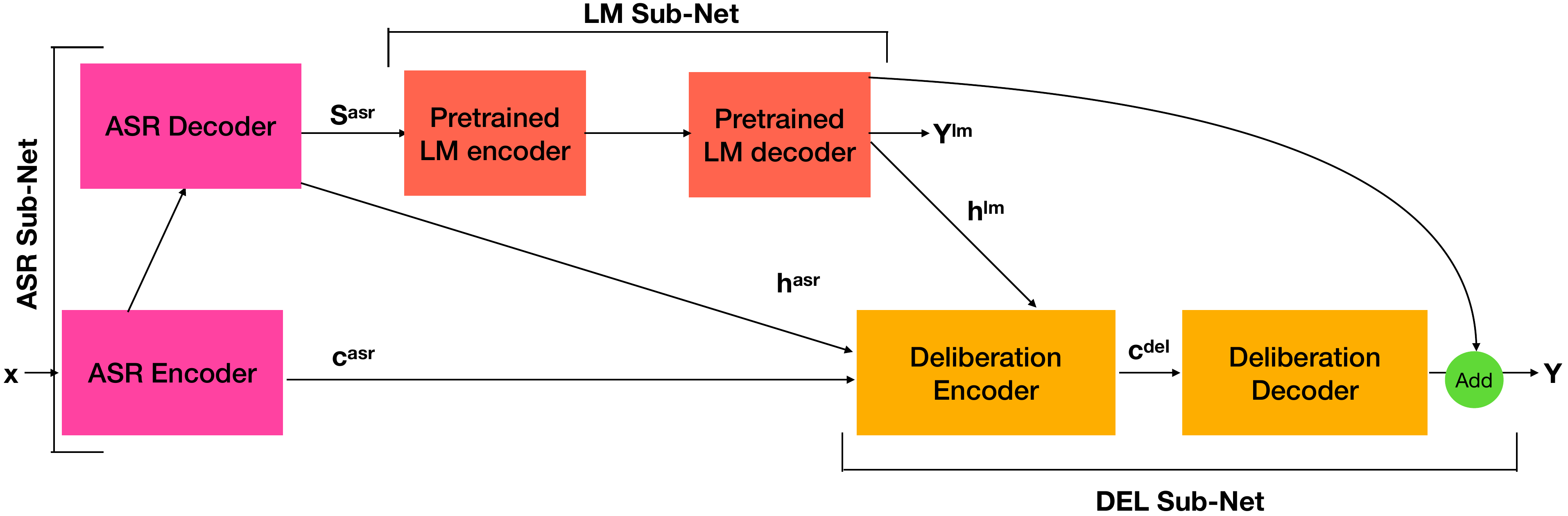}
\caption{Schematics of our three-pass SLU system that incorporates pretrained ASR and LM.}
\label{fig:system-overview}
\vskip -0.1in
\end{figure*}

\section{Introduction}
\input{sections/introduction.tex}

\section{Problem Formulation}
The task of sequence labeling has been extensively studied in NLU systems, which take a text sequence $S$ of length $N$ and vocabulary $\mathcal{V}$ as input, represented as $S = \{s_n \in \mathcal{V} | n=1,\dots , N\}$. The model generates an output sequence $Y = \{ y_o \in \mathcal{L} | o=1,\dots ,O \}$, where $\mathcal{L}$ is the label set and $O$ is the length of the output sequence. NLU systems, usually based on pretrained LMs, use the maximum a posteriori (MAP) theory to maximize the posterior distribution $P(Y|S)$.

In contrast, spoken language understanding (SLU) systems aim to generate the output sequence directly from the speech feature sequence $X = \{ \mathbf{x}_t | t=1,\dots ,T \}$ where $T$ is the length of this sequence. SLU models use the MAP theory to output $\hat{Y}$ such that $\hat{Y}={\text{argmax}} ~P(Y|X)$. We describe how we model this posterior distribution in Sec.~\ref{sec: method} below.

\section{Our three-pass SLU model}
\label{sec: method}
Our method aims to incorporate both pretrained ASR and LMs inside the SLU framework. We assume a pretrained ASR model that computes ASR transcript $S^{\text{asr}}$ from spoken sequence $X$ by maximizing $P(S|X)$ and pretrained LM that seeks to estimate label sequence $Y^{\text{lm}}$ from text sequence $S$ by maximizing $P(Y|S)$. By regarding the $S^\text{asr}$ and $Y^\text{lm}$ as a probabilistic variable, we can theoretically incorporate the ASR and LM modules into our formulation of the posterior distribution as shown:
\begin{equation}
P(Y|X) = \sum_{S^{\text{asr}}} \sum_{Y^{\text{lm}}} P(Y|X,S^{\text{asr}},Y^{\text{lm}})P(Y^{\text{lm}}|X,S^{\text{asr}})P(S^{\text{asr}}|X)
\label{eq:theory}
\end{equation}
 We use Viterbi approximation to make the above computation tractable.
We further assume the conditional independence of $Y^{\text{lm}}|S^{\text{asr}}$ from X to simplify Eq. ~\ref{eq:theory}:
\begin{align}
    P(Y|X) &\approx \max_{S^{\text{asr}}}\max_{Y^{\text{lm}}} \underset{\text{DEL Sub-Net}}{\underbrace{P(Y|X,S^{\text{asr}},Y^{\text{lm}})}}\underset{\text{LM Sub-Net}}{\underbrace{P(Y^{\text{lm}}|S^{\text{asr}})}}\underset{\text{ASR Sub-Net}}{\underbrace{P(S^{\text{asr}}|X)}} \label{eq:final_formulation}
\end{align}
To achieve the formulation described in Eq.~\ref{eq:final_formulation}, this work proposes a three-pass SLU architecture, which is illustrated in Figure~\ref{fig:system-overview}. The architecture comprises three subnetworks: ASR, LM, and deliberation. In the first pass, the ASR subnetwork models $P(S^{\text{asr}}|X)$, as detailed in Sec.~\ref{sec:asr_sub_net}. The output of this pass, $S^{\text{asr}}$, serves as the input to the LM subnetwork in the second pass, which models $P(Y^{\text{lm}}|S^{\text{asr}})$ and generates $Y^{\text{lm}}$, as discussed in Sec.~\ref{sec:lm_sub_net}. Finally, in the third pass, the deliberation subnetwork uses the outputs of the first and second passes to model ${P(Y|X,S^{\text{asr}},Y^{\text{lm}})}$ and make the final predictions, as explained in Sec.~\ref{sec:del_sub_net}.
\subsection{ASR subnetwork  $P(S^\text{asr}|X)$}
\label{sec:asr_sub_net}
The input speech $X$ is first passed through ASR encoder ($\text{Encoder}^{\text{asr}}$) to produce acoustic embeddings ($\textbf{c}^{\text{asr}}$) as shown:
\begin{equation}
    \textbf{c}^{\text{asr}}= \text{Encoder}^{\text{asr}}(X)\label{aco_eq}
\end{equation}
This acoustic embedding is a sequence of latent representations $\textbf{c}^{\text{asr}}=(\textbf{c}^{\text{asr}}_{1},\textbf{c}^{\text{asr}}_{t},..,\textbf{c}^{\text{asr}}_{T})$.
The ASR decoder ($\text{Decoder}^{\text{asr}}$) then maps the acoustic embedding $\textbf{c}^{\text{asr}}$ and preceding ASR tokens generated by the ASR decoder $s^{\text{asr}}_{1:n-1}$ to ASR decoder hidden representations $\textbf{h}_{n}^{\text{asr}}$.
\begin{equation}
\textbf{h}_{n}^{\text{asr}} = \text{Decoder}^{\text{asr}}(\textbf{c}^{\text{asr}}_{1:T}, s^{\text{asr}}_{1:n-1}) \label{asr_decoder_eq}
\end{equation}
The likelihood of each token in the ASR transcript is given by $\text{Out}^{\text{asr}}$ which denotes a linear layer that maps decoder output $\mathbf{h}_n^{\text{asr}}$ to vocabulary $\mathcal{V}$ followed by softmax function.
\begin{equation}
P(s^{\text{asr}}_n | X, s^{\text{asr}}_{1:n-1}) = \text{Softmax}(\text{Out}^{\text{asr}}(\mathbf{h}_n^{\text{asr}})) \label{intermediate_eq}
\end{equation}
The likelihood of the entire ASR transcript is computed by composing the conditional probabilities at each step for N tokens.
\begin{equation}
P(S^{asr}|X) = \prod_{n=1}^{N} P(s^{\text{asr}}_n | X, s^{\text{asr}}_{1:n-1})\label{ASR_prob_eq}
\end{equation}
\subsection{LM subnetwork  $P(Y^{\text{lm}}|S^{\text{asr}})$}
\label{sec:lm_sub_net}
The ASR transcript is then provided as input to the encoder of pretrained LM ($\text{Encoder}^{\text{lm}}$) as shown below.
\begin{equation}
    \textbf{c}^{\text{lm}}= \text{Encoder}^{\text{lm}}(S^{\text{asr}})
\end{equation}
Similar to the ASR decoder, the LM decoder ($\text{Decoder}^{\text{lm}}$) then maps the semantic embedding $\textbf{c}^{\text{lm}}$ to hidden LM decoder representations $\textbf{h}^{\text{lm}}$ as shown below:
\begin{equation}
    \textbf{h}_{o}^{\text{lm}} = \text{Decoder}^{\text{lm}}(\textbf{c}^{\text{lm}}_{1:N}, y^{\text{lm}}_{1:o-1}) \label{LM_dec_rep_eq}
\end{equation}
The likelihood of the entire SLU label sequence is computed similarly to Eq.~\ref{ASR_prob_eq} as shown below 
\begin{equation}
P(y_o^{\text{lm}}  | S^{\text{asr}}, y^{\text{lm}}_{1:o-1}) = \text{Softmax}(\text{Out}^{\text{lm}}(\mathbf{h}_o^{\text{lm}}))\label{LM_decoder_eq}
\end{equation}
\begin{equation}
P(Y^{\text{lm}}|S^{\text{asr}}) = \prod_{o=1}^{O} P(y_o^{\text{lm}} | S^{\text{asr}}, y^{\text{lm}}_{1:o-1})\label{LM_prob_eq}
\end{equation}
\subsection{Deliberation subnetwork  $P(Y|X,S^{\text{asr}},Y^{\text{lm}})$}
\label{sec:del_sub_net}
Finally, we formulate $P(Y|X,S^{\text{asr}},Y^{\text{lm}})$ by conditioning on $X$ using ASR encoder representation $\textbf{c}^{\text{asr}}$ (Eq.~\ref{aco_eq}), $S^{\text{asr}}$ using ASR decoder representations $\textbf{h}^{\text{asr}}$ (Eq.~\ref{asr_decoder_eq}) and $Y^{\text{lm}}$ using LM decoder representations $\textbf{h}^{\text{lm}}$ (Eq.~\ref{LM_dec_rep_eq}) in the following two ways.
\subsubsection{Concatenation Integration}
\label{sec:concate_int}
We ensure that $\textbf{h}^{\text{lm}}$ has the same embedding dimension as acoustic embedding $\textbf{c}^{\text{asr}}$ by passing it through a linear layer. The ASR encoder, ASR decoder, and LM decoder representations are concatenatenated
to produce a sequence of length $K$ where $K=T+N+O$. This sequence is fed as input to a deliberation encoder $\text{Encoder}^{\text{del}}$ to produce joint embedding ($\textbf{c}^{\text{del}}$):
\begin{equation}
    \textbf{c}^{\text{del}} = \text{Encoder}^{\text{del}}(\textbf{c}^{\text{asr}}||\textbf{h}^{\text{asr}}|| \textbf{h}^{\text{lm}})
\label{deliberation_enc}
\end{equation}
Similar to Eq.~\ref{LM_decoder_eq}, the deliberation decoder ($\text{Decoder}$) maps this joint embedding $\textbf{c}^{\text{del}}$ and previous tokens output by the decoder $y_{1:o-1}$ to generate SLU output sequence $Y$.
\begin{equation}
    \textbf{h}_{o}= \text{Decoder}(\textbf{c}^{\text{del}}_{1:K},y_{1:o-1})
\end{equation}
\begin{equation}
    P(y_o | X, S^{\text{asr}}, Y^{\text{lm}}, y_{1:o-1}) = \text{Softmax}(\text{Out}(\mathbf{h}_o))\label{second_pass_decoder}
\end{equation}
Our architecture further adds a residual connection to posteriors obtained from pretrained LM (Eq.~\ref{LM_decoder_eq}) and performs a weighted combination of the 2 posteriors to make the final prediction.
\begin{equation}
    \hat{h}_o= (\alpha*\text{Out}(\mathbf{h}_o)+(1-\alpha)*\text{Out}^\text{lm}(\mathbf{h}_o^{lm}))\label{residual_connection_combination}
\end{equation}
\begin{equation}
    P(y_o | X, S^{\text{asr}}, Y^{\text{lm}}, Y_{1:o-1}) = \text{Softmax}(\hat{h}_o)\label{residual_connection}
\end{equation}
where $\alpha$ is a learnable parameter. Note that 3-pass SLU model is trained using LM vocabulary to facilitate this combination.

\subsubsection{Cross Attention Integration}
\label{sec:cross_att_int}
Inspired by prior work on compositional speech processing models~\cite{dalmia2021searchable,Compositional_E2E}, we also experiment with incorporating acoustic embedding $\textbf{c}^{\text{asr}}$ using the multi-sequence cross attention in our deliberation decoder. In the formulation, the ASR decoder and LM representations are concatenated together and then a joint embedding is produced by the deliberation encoder:
\begin{equation}
    \textbf{c}^{\text{del}} = \text{Encoder}^{\text{del}}(\textbf{h}^{\text{asr}}|| \textbf{h}^{\text{lm}})
\label{deliberation_enc_2}
\end{equation}
The deliberation decoder then conditions on acoustic embedding $\textbf{c}^{\text{asr}}$ and joint embedding $\textbf{c}^{\text{del}}$ to make the final prediction:
\begin{equation}
    \textbf{h}_{o}= \text{Decoder}(\textbf{c}^{\text{del}},\textbf{c}^{\text{asr}},y^{1:o-1})
\label{decode_cross}
\end{equation}
The final prediction is similarly made by using a weighted combination with LM posteriors as shown in Eq.~\ref{residual_connection_combination}.

\subsection{Training and Inference}
\label{sec:training_details}
Similar to prior work on 2-pass SLU systems~\cite{two_pass_slu}, we use a three step training process for our SLU system. In the first step, only the ASR subnetwork is trained using Eq.~\ref{ASR_prob_eq}. In the second step, we train both the LM encoder and decoder with ground truth ASR transcript using Eq.~\ref{LM_prob_eq}. In the third step, the deliberation encoder ($\text{Encoder}^{\text{del}}$) and decoder ($\text{Decoder}$) are trained using Eq.~\ref{residual_connection} while using the pretrained ASR and LM subnetwork from step 1 and step 2. We also tried fine-tuning the entire network in step 3 but our initial experiments did not show any significant performance gains. We use teacher forcing on LM and deliberation decoder and also experiment with and without teacher forcing of ASR transcripts during step 3 in Sec.~\ref{sec:teacher_forcing}.

Our decoding process based on Eq.~\ref{eq:final_formulation} also consists of three steps. In the first step, the ASR subnetwork generates the ASR transcript by maximizing $P(S^\text{asr}|X)$ (Eq.~\ref{ASR_prob_eq}). In the second step, the LM subnetwork obtains the initial SLU predictions $Y^\text{lm}$ by maximizing $P(Y^\text{lm}|S^\text{asr})$ (Eq.~\ref{LM_prob_eq}). The deliberation decoder then generates the final SLU predictions, $Y$, by maximizing $P(Y|X,S^{\text{asr}},Y^{\text{lm}})$ (Eq.~\ref{residual_connection}).
Beam search is used at all 3 steps of inference which also helps
improve ASR ($\textbf{h}^{\text{asr}}$) and LM ($\textbf{h}^{\text{lm}}$) decoder representations.

\section{Experiments}
\label{sec: problem formulation}
\subsection{Datasets}
To demonstrate the effectiveness of our 3-pass SLU model, we conducted experiments on two publicly available spoken named entity recognition datasets, namely SLURP~\cite{SLURP} and SLUE-VoxPopuli~\cite{SLUE}. The SLURP corpus consists of single-turn user conversations with a home assistant, making it more relevant to commercial SLU applications. To ensure consistency with prior work~\cite{SLURP,ESPnet-SLU}, we augment our training set with synthetic data. The corpus comprises 40.2 hours of audio, with 11,514 utterances in train set, 6.9 hours with 2,033 utterances in development set, and 10.3 hours with 2,974 utterances in test set.

The recently released SLUE benchmark is a publicly available, naturally produced speech dataset that contrasts with the read speech data used in most SLU benchmarks. We are specifically interested in the SLUE-VoxPopuli dataset \cite{SLUE}, which contains NER annotations and consists of European Parliament recordings. The corpus comprises 5,000 utterances, with 14.5 hours of audio in the train set and 1,753 utterances, containing nearly 5 hours of audio in the development set. The test sets released for SLUE are blind, without ground truth labels; therefore, we compare different methods using the development set.

We report performance on SLURP using the SLU F1~\cite{SLURP} which weights the match of an entity label with the character and word error rate of the entity mention. SLUE is evaluated using F1 which exactly matches both the entity label and entity mention. We report micro-averaged F1 for all our experiments.

\subsection{Baseline}
We compare our proposed 3-pass SLU model with cascaded and other E2E SLU systems. We also train Cascaded SLU model with BART-large as NLU which is essentially the same as the first 2 passes of our 3 pass SLU model. We also report performance of a compositional E2E SLU model that predicts the SLU label sequence using a decoder (referred to as “Compositional E2E SLU with Direct E2E formulation” in~\cite{Compositional_E2E}\footnote{Note that we did not compare with token classification models, as our focus was on sequence generation for SLU tasks.}). We also compare our approach with the 2-pass SLU~\cite{two_pass_slu} model, where the second pass model can utilize semantic representations from a pretrained LM, along with acoustic representations.

\subsection{Experimental Setups}
Our models are implemented in pytorch \cite{pytorch}
and the experiments are conducted through the ESPNet-SLU \cite{ESPnet-SLU} toolkit. Following the approach in \cite{Compositional_E2E}, we use a 12-layer conformer block for the encoder and a 6-layer transformer block for the decoder in our ASR model for both datasets. Each attention block consists of 8 attention heads, a dropout rate of 0.1, an output dimension of 512, and a feedforward dimension of 2048 for the SLURP dataset. For the SLUE dataset, each attention block has 4 attention heads, a dropout rate of 0.1, an output dimension of 256 for the encoder and 2048 for the decoder, and a feedforward dimension of 1024 for the encoder and 2048 for the decoder. Our ASR models achieved a word error rate (WER) of 30.4 for the SLUE dataset and 16.3 for the SLURP dataset. We train our ASR models to generate BPE tokens with a size of 500 for the SLURP dataset and 1000 for the SLUE dataset.

We use BART large \cite{BART} as our pretrained LM and train it with the HuggingFace Seq2Seq Trainer \cite{wolf-etal-2020-transformers}. We train the models with a learning rate of 1e-5 and 5k warmup steps. Our 3-pass SLU architecture uses the above-described ASR and LM as the ASR and LM subnetworks, respectively. For our 3-pass SLU architecture with concatenation integration, our deliberation encoder consists of a 4-layer conformer block with a feedforward dimension of 2048 for both datasets. For our 3-pass SLU architecture with cross attention integration, our deliberation encoder consists of a 6-layer transformer block with a feedforward dimension of 2048 for both datasets. The attention heads, output dimension, and other parameters of the deliberation encoder are the same as the ASR encoder of the respective datasets. The third-pass decoder for both datasets has the same architecture as the ASR decoder giving a total parameter size of 618 M for SLURP and 482 M for SLUE dataset. We train our 3-pass SLU models to generate BART large vocabulary tokens to facilitate combination with posteriors from the LM subnetwork (Eq.~\ref{residual_connection_combination}).

We also train 2-pass SLU model \cite{two_pass_slu} using the same encoder and decoder as our ASR model and the same deliberation encoder as our 3-pass SLU model w/ concatenation integration. 
We perform SpecAugment \cite{specaugment} for data augmentation and apply dropout \cite{dropout} and label smoothing \cite{label-smoothing}.
Models were trained using 4 NVIDIA A40 GPUs. All model, training and inference parameters were selected based on validation performance.

\begin{table}[t]
  \centering
    \resizebox {0.82\linewidth} {!} {
\begin{tabular}{lcc}
\toprule
& \multicolumn{1}{c}{SLURP} & \multicolumn{1}{c}{SLUE}\\ 
\cmidrule(r){2-3}
Model & SLU F1 $\uparrow$ & F1 $\uparrow$ \\ 
 \midrule
Casacaded SLU \cite{Compositional_E2E} & 73.3 & 48.6  \\
Casacaded SLU with BART (Ours) & 77.9 & 55.8 \\ 
Direct E2E SLU \cite{Compositional_E2E} & 77.1 & 54.7\\ 
Compositional E2E SLU \cite{Compositional_E2E} & 77.2 & 50.0 \\
2-pass E2E SLU \cite{two_pass_slu}  & 77.4 & 52.2 \\
\midrule
3-pass SLU  \\
w/ Concatenation \Sref{sec:concate_int} & 78.4 & 56.7\\
w/ Cross attention \Sref{sec:cross_att_int} & \textbf{78.5} & \textbf{57.2} \\
\bottomrule
\end{tabular}
}
\caption{Results presenting performance of 3-pass SLU system.} 
\vskip -0.2in
\label{tab:main-results}
\end{table}
\begin{table}[t]
  \centering
  \resizebox {0.82\linewidth} {!} {
\begin{tabular}{lcc}
\toprule
& \multicolumn{1}{c}{SLURP} & \multicolumn{1}{c}{SLUE} \\ 
\cmidrule(r){2-3}
Model & SLU F1 $\uparrow$ & F1 $\uparrow$ \\ 
\midrule
3-pass SLU   w/ Cross attention & 78.5  & 57.2 \\
w/o teacher forcing & 78.4 & 54.1\\
w/o $h^{\text{asr}}$ in Eq.~\ref{deliberation_enc_2} & 78.4 & 56.7\\
w/o $c^{\text{asr}}$ in  in Eq.~\ref{decode_cross} & 78.4 & 56.6\\
w/o $h^{\text{lm}}$ in Eq.~\ref{deliberation_enc_2} & 74.5 & 56.3\\
w/o Residual connection in Eq.~\ref{residual_connection} & 77.8 & 50.9\\
\bottomrule
\end{tabular}
}
\caption{Results presenting ablation of our 3-pass SLU system.}
\vskip -0.3in
\label{tab:teacher-force-results}
\end{table}

\subsection{Results and Discussion}
Table~\ref{tab:main-results} shows that our proposed 3-pass SLU model outperforms both cascaded and E2E SLU models on both datasets. It also performs significantly better than compositional E2E SLU models. Our 3-pass SLU models are a more effective way of utilizing pretrained LMs than the previously proposed 2-pass SLU model. Our experiments show that cross attention integration (Eq.~\ref{decode_cross}) is a better way of conditioning on ASR and LM output than concatenation integration (Eq.~\ref{deliberation_enc}). Our performance gains over cascaded systems are more significant for the SLUE dataset, which consists of natural conversation speech instead of read speech and is therefore more acoustically challenging. Hence our approach of E2E integration of ASR and LM subnetworks helps avoid cascading errors from the ASR transcript.
\subsubsection{Ablation study}
\label{sec:teacher_forcing}
We conduct ablation experiments on our 3-pass SLU model to gain a better understanding of the relative importance of different formulation decisions. Specifically, we first perform an ablation study in which we used ASR hypotheses instead of ground truth transcripts during the third step of training (see Sec.~\ref{sec:training_details}). Table~\ref{tab:teacher-force-results} demonstrates that teacher forcing enhances the SLU performance on both datasets. The impact of teacher forcing is more substantial on the SLUE dataset since it has a higher WER, and errors in ASR transcripts can pose a challenge to the model's training. We also experiment with masking out representations of the ASR encoder $c^{\text{asr}}$ and ASR decoder $h^{\text{asr}}$ from Eq.\ref{decode_cross} and Eq.\ref{deliberation_enc_2}, respectively, and observe a slight drop in performance. Masking out the LM decoder representation $h^{\text{lm}}$ from Eq.\ref{deliberation_enc_2} has a more significant negative impact, highlighting its importance. The exclusion of a residual connection (Eq.~\ref{residual_connection}) to posteriors from a pretrained LM also results in significant performance degradation. We conclude that our formulation decisions as described in Sec.~\ref{sec: method} achieve best performance.
\begin{table}[t]
\resizebox {\linewidth} {!} {
  \centering
\begin{tabular}{lcc|cc|c}
\toprule
& \multicolumn{2}{c|}{Cascaded SLU} & \multicolumn{2}{c}{3 pass SLU} \\ 
\cmidrule(r){2-5}
WER & SLU F1 $\uparrow$ & Label F1 $\uparrow$ & SLU F1 $\uparrow$ & Label F1 $\uparrow$ & \# Utterances \\ 
0.0 & 95.8 & 96.6 & 95.7 & 96.5 & 7164\\
(0.0,0.25] & 75.2 & 85.7 & 76.3 & 86.3 & 2847\\
(0.25,1.0] & 51.1 & 62.1 & 51.9 & 62.7 & 3067 \\
\bottomrule
\end{tabular}
}
\caption{Results comparing performance of Cascaded model using BART large \& 3-pass model with cross attention integration across different ASR difficulties on SLURP dataset.} 
\label{tab:werr-results}
\vskip -0.3in
\end{table}
\subsubsection{Performance based on ASR WER}
To assess the efficacy of our proposed 3-pass SLU model, we compare its performance with that of a cascaded model across utterances with varying levels of acoustic complexity on SLURP dataset. As in prior work~\cite{ASR_Pretrain1}, we also divide our test set into groups based on acoustic complexity, as quantified by the WER of ASR transcripts, with each group containing a similar number of utterances. Our findings in Table~\ref{tab:werr-results} indicate that both models perform similar when the WER is perfect, and that most of the performance gains are observed for utterances with errors in the ASR transcript. Further the maximum gain in performance is seen for utterances with WER between 0 and 0.25.

We also evaluate using Label-F1 metric, which only considers the matches of entity labels. The results show that improvement is smaller for Label-F1, suggesting that a part of the improvement is due to correctly identifying entity mentions, even when there
are errors in the ASR transcript.

\section{Conclusion}
\label{sec: conclusion}
We present a novel 3-pass SLU model that integrates ASR and LM sub-networks into the E2E SLU formulation. Our study demonstrates that our model outperforms both cascaded and other E2E SLU models on sequence generation tasks, such as named entity recognition. 
We conduct an ablation study to evaluate the relative contributions of our design choices. 
Additionally, we observe that the majority of performance improvements from our model are seen on utterances with inaccurate ASR output. In future work, we plan to extend our proposed architecture to other speech processing tasks, such as speech translation.

\section{Acknowledgements}
\label{sec:acknowledgements}
This work used PSC Bridges2 and NCSA Delta through allocation CIS210014 from the Advanced Cyberinfrastructure Coordination Ecosystem: Services \& Support (ACCESS) program,
which is supported by National Science Foundation grants
\#2138259, \#2138286, \#2138307, \#2137603, and \#2138296.
\section{References}
{
\printbibliography
}
\end{document}

%% file: sections/introduction.tex
Spoken Language Understanding (SLU) involves extracting semantic meaning or linguistic structure from spoken utterances. It has a wide range of applications in commercial devices like chatbots and voice assistants~\cite{socialbot, snips-voice-platform}. To advance research in this field, new tasks and benchmarks~\cite{STOP,SLUE2} have been proposed. Recently, there has been interest in performing sequence labeling (SL) tasks, such as entity recognition~\cite{SLURP,SLUE} and semantic parsing~\cite{STOP}, directly from spoken utterances. SL systems aim to tag each token in an utterance to provide insights into its linguistic structure and semantic meaning~\cite{Jurafsky}. When performing SL directly on spoken utterances, there is an additional complexity of recognizing the mention of the entity tag~\cite{Compositional_E2E,zhai2004using}.

Conventional SLU systems~\cite{palmer-ostendorf-2001-improving,Horlock2003DiscriminativeMF,BECHET2004207} employ a cascaded approach for sequence labeling, where an automatic speech recognition (ASR) system first recognizes the spoken words from the input audio and a natural language understanding (NLU) system then tags the words in the predicted text. These cascaded approaches can effectively utilize pretrained ASR and NLU systems. However, they suffer from error propagation~\cite{Tran_SLU} as errors in the ASR transcripts can adversely affect downstream SLU performance. As a result, these systems particularly struggle in noisy or adverse scenarios where we do not have a good ASR system. These scenarios are particularly common for spontaneous speech~\cite{SWBD,SLUE}, for instance, in spoken conversation scenarios. Consequently, in this work, we focus on end-to-end (E2E) SLU systems. E2E SLU systems~\cite{ESPnet-SLU,end-to-end1} aim to predict entity tags and mentions directly from speech. These E2E SLU systems can avoid the cascading of errors but cannot directly utilize strong acoustic and semantic representations from pretrained ASR systems and language models (LMs) due to a vocabulary mismatch between pretrained ASR and LMs~\cite{SLUE}.

Most SLU datasets have little labeled data since they are expensive and time-consuming to collect. To address this challenge, there has been a significant amount of interest in using pretrained ASR and LMs in E2E SLU architecture. Prior works~\cite{ASR_Pretrain1,SLUE,ESPnet-SLU,ASR_Pretrain2,ASR_Pretrain3,lai2021semi,ASR_Pretrain4,ASR_Pretrain5} have explored either pretraining SLU models on larger ASR corpora or using the encoder of a pretrained ASR model as a frontend. Recently, compositional E2E SLU~\cite{Compositional_E2E} models have been proposed which show better adaptability with pretrained ASR systems, but they are still ineffective at utilizing pretrained NLU models. Similar efforts~\cite{agrawal2020tie,speechbert,chung2020splat,LM_Pretrain1,LM_Pretrain2,LM_Pretrain3} have been made to incorporate LMs within the SLU framework. One approach~\cite{chung2020splat,huang2020leveraging} aligns the acoustic embedding of the SLU model with text embeddings produced by the LM encoder. Another approach~\cite{seo2021integration} fuses the ASR and LM encoders using a linear layer called an interface. Other works~\cite{two_pass_slu,lai2021semi,Ddel1} concatenate the acoustic embeddings and semantic embeddings generated from a pretrained LM to make the final prediction using a deliberation network~\cite{xia2017deliberation,hu2020deliberation}.

In this study, we draw inspiration from previous work on 2-pass SLU~\cite{two_pass_slu} models and introduce a formulation that integrates ASR and LM subnetwork inside the SLU framework using a probabilistic approach. In the first pass, the ASR subnetwork predicts the ASR transcript from the input speech, followed by the LM subnetwork, which inputs the ASR transcript to predict the SLU label sequence. In the third pass, our formulation incorporates the ASR encoder embedding, ASR decoder embedding, and pretrained LM decoder embedding using an encoder-decoder architecture. Thus, we achieve modular-based end-to-end SLU system by maintaining the ability of each pretrained subnetwork. This not only helps both the subnetworks to be trained using their own vocabulary but also leads to better generation ability of LM as it's closer to it's NLU formulation. Further, by conditioning on ASR and LM representation during deliberation, it can recover from errors in ASR transcript, helping to outperform existing cascaded systems.

The key contributions of our work are:
\begin{itemize}
    \item Integrating ASR and LM networks inside SLU formulation using probabilistic approach.
    \item Proposing a novel 3-pass SLU model that can incorporate pretrained ASR and LM for the sequence generation task.
    \item Demonstrating that our proposed model can improve the SLU performance for named entity recognition on two benchmark datasets, namely SLURP~\cite{SLURP} and SLUE-VoxPopuli~\cite{SLUE}.
    \item Conducting an analysis that reveals that our improvements are most pronounced on acoustically challenging utterances.
\end{itemize}